\documentclass[a4paper,fleqn]{cas-dc}

\usepackage[numbers]{natbib}
\usepackage{lineno,hyperref}
\usepackage{amsmath}
\usepackage{amssymb}
\usepackage{subfigure}
\DeclareMathOperator*{\argmin}{argmin}
\usepackage{multirow}
\usepackage{color}
\usepackage{float}
\usepackage{hyperref}

\def\tsc#1{\csdef{#1}{\textsc{\lowercase{#1}}\xspace}}
\tsc{WGM}
\tsc{QE}
\tsc{EP}
\tsc{PMS}
\tsc{BEC}
\tsc{DE}

\emergencystretch=\maxdimen
\begin{document}
\shorttitle{Structure-Enhanced Meta-Learning For Few-Shot Graph Classification}
\shortauthors{Shunyu Jiang et~al.}

\title [mode = title]{Structure-Enhanced Meta-Learning For Few-Shot Graph Classification}                      


\author[1]{Shunyu Jiang}[style=chinese]

\author[2]{Fuli Feng}[style=chinese]

\cormark[1]
\cortext[cor1]{Corresponding author. National University of Singapore, Singapore. $\textit{Email address:}$ fulifeng93@gmail.com}

\author[1]{Weijian Chen}[style=chinese]
\author[3]{Xiang Li}[style=chinese]
\author[1]{Xiangnan He}[style=chinese]



\address[1]{University of Science and Technology of China, Hefei,Anhui, 230027, China }
\address[2]{National University of Singapore, 119077, Singapore}
\address[3]{University of Hong Kong}

\begin{abstract}
Graph classification is a highly impactful task that plays a crucial role in a myriad of real-world applications such as molecular property prediction and protein function prediction.
Aiming to handle the new classes with limited labeled graphs, few-shot graph classification has become a bridge of existing graph classification solutions and practical usage.
This work explores the potential of metric-based meta-learning for solving few-shot graph classification.
We highlight the importance of considering structural characteristics in the solution and propose a novel framework which explicitly considers \textit{global structure} and \textit{local structure} of the input graph. 
An implementation upon GIN, named SMF-GIN, is tested on two datasets, Chembl and TRIANGLES, where extensive experiments validate the effectiveness of the proposed method. 
The Chembl is constructed to fill in the gap of lacking large-scale benchmark for few-shot graph classification evaluation, which is released together with the implementation of SMF-GIN at: \url{https://github.com/jiangshunyu/SMF-GIN}.

\end{abstract}


\maketitle

\section{Introduction}

Graphs are widely used to represent relation data in plenty of real-world applications such as social networks~\citep{dgi}, chemical molecules~\citep{pre-GNN}, and biological proteins~\citep{graphsage}, due to their strong capability of modeling relationships and structure.
A surge of attention has been focused on graph-based analytical tasks, including graph classification such as predicting the chemical properties of molecules~\citep{pre-GNN}. 
The existing graph classification solutions largely follow supervised learning where sufficient labeled graphs are required for each class~\citep{GIN}.
Despite the success on well-labeled classes, many real-world scenarios are faced with few-shot classes where only limited labeled graphs are available due to the high cost of labeling~\citep{supclass} or the occurrence of new classes~\citep{few-node}. 
For instance, in chemical-pharmaceutical industry, the requirement of predicting new molecule properties keeps occurring, e.g., predicting the inhibition against COVID-19, where the labeled molecules are very limited.
To address this dilemma, it is imperative to shift the attention to few-shot graph classification.

As a few-shot classification task, a natural way is extending existing meta-learning solutions for conventional few-shot tasks such as image classification~\citep{few-shot} to graphs by additionally considering the graph structure.
Indeed, the vanilla model-agnostic gradient-based meta-learning (MAML)~\citep{MAML} has been extended to graph classification~\citep{maml-GNN} by utilizing graph neural network (GNN) as the back-bone model.
The key idea is to learn a GNN initialization that can be well generalized and fast adapted to new classes with the generalization performance as the supervision, which simply counts on the GNN to encode graph structure.
In addition to the prevalence of gradient-based methods like MAML in meta-learning,
there are also works that demonstrate the effectiveness of metric-based meta-learning~\citep{siamese} in conventional few-shot classification. Its target is to learn sample representation and distance measure that can be generalized to the new classes where the prediction is made according to the distance to the labeled samples.
Along this line,~\citeauthor{supclass} propose a graph distance measure in the spectral domain, which validates the merit of considering graph structure.
However, the measure in spectral domain might not well utilize the graph structure, making it inflexible and hard to be adjusted according to the graph properties. 
Therefore, it is worthwhile exploring meta-learning for few-shot graph classification with \textbf{explicit consideration of graph structure}, which has received relatively little scrutiny.

\begin{figure}[t]
	\centering
	\includegraphics[width=0.4\textwidth]{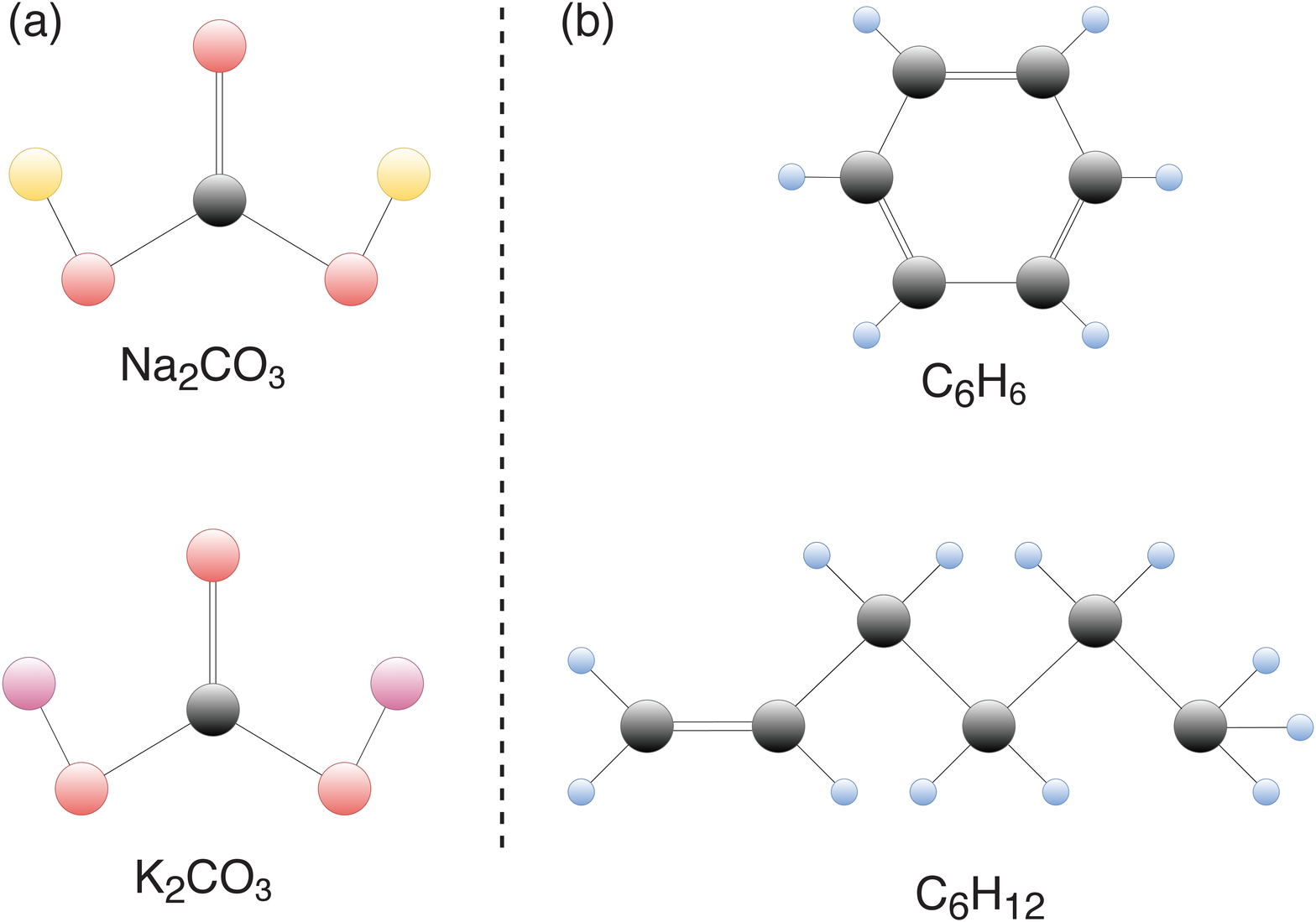}
	\vspace{-5pt}
	\caption{An illustration of molecule graph.}
	\vspace{-0.5cm}
	\label{fig:molecule}
\end{figure}

We argue that the structural information of graph data consists of two parts, namely 
\textit{local structure} and \textit{global structure}. 
On the one hand, some attributes of a graph depend on the substructure of the graph, that is, the \textit{local structure} plays a decisive role in predicting the graph's attribute class. 
For example, as shown in Figure~\ref{fig:molecule}(a), the structures of $Na_{2}CO_{3}$ and $K_{2}CO_{3}$ both contain carbonate groups (i.e., $CO_3$), making them exhibit similar characteristics such as the weak alkalinity when dissolved in water.
As such, the \textit{local structure} should be well represented when measuring the distance of graphs. 
On the other hand, as GNN has become a default choice for extracting graph representations, the \textit{global structure} of a graph also affects whether the distance calculated from the representations can be well generalized. 
As shown in Figure~\ref{fig:molecule}(b), the graph representation at deep layers will be over-smoothing for $C_6H_6$ with small diameters~\citep{oversmoothing}, while for $C_6H_{12}$ with large diameters, the graph representation at shallow layer will be insufficient for describing the overall graph~\citep{deep-gcn}.
As the graphs in new classes may have distinct \textit{global structure} and \textit{local structure}, it is even more challenging to model them both effectively.

In this work, we propose a metric-based meta-learning framework for few-shot graph classification that explicitly considers the structure characteristics of graph data when calculating graph representation.
In this framework, a multi-layer GNN is employed to encode the input graph where different layers capture the information at different granularities. 
To better account for the \textit{global structure}, instead of simply aggregating (e.g., concatenating) the representations from different GNN layers, we devise an adaptive aggregation module that assigns weights for different layers in a graph-specific manner. 
As to the \textit{local structure}, the framework further extracts the representations of representative substructures (e.g., the scaffold~\citep{MoleculeNet} of molecular)
which are adaptively fused into the overall graph representation. 
In this way, the framework can achieve more accurate distance metric to enhance the graph classification on new classes. 
Moreover, the framework is transparent, shedding light on how the \textit{global structure} and \textit{local structure} effect the classification.

In particular, we implement the proposed framework over a state-of-the-art model for graph classification, Graph Isomorphism Network (GIN)~\citep{GIN}. 
We evaluate our solution, named SMF-GIN, over the existing datasets of few-shot graph classification such as TRIANGLES~\citep{supclass}.
However, the existing datasets not only contain few classes but are also small in size, making it difficult to truly validate the performance of the model.
To facilitate the research and evaluation of the few-shot graph classification task, we extract a large-scale multi-class graph classification dataset in the Chembl chemical molecule database~\citep{MoleculeNet}, named Chembl, which is believed to be a general benchmark dataset.
Extensive experiments on TRIANGLES and Chembl validate the effectiveness and rationality of the proposed method.

The main contributions of our work are as follows:
\begin{itemize}[leftmargin=*]
\item Proposing a metric-based meta-learning solution for the few-shot graph classification task, which explicitly encodes the \textit{local structure} and \textit{global structure} of a graph.
\item Constructing a general large-scale multi-class benchmark dataset to facilitate future research on few-shot graph classification.
\item Conducting extensive experiments on two datasets to justify the effectiveness of our design.
\end{itemize}

\section{Method}
In this section, we first describe the problem setup (Section~\ref{ssec:problem}) followed by the introduction of the proposed framework (Section~\ref{ssec:framework}).
Lastly, we detail the consideration of graph \textit{local structure} and \textit{global structure}.

\subsection{Problem Setup}\label{ssec:problem}
First of all, we define the problem of the few-shot graph classification, where a set of graphs
$\{\mathcal{G}_{1},\mathcal{G}_{2},\cdots,\mathcal{G}_{m}\}$ and their labels $\{\mathbf{y}_{1},\mathbf{y}_{2},\cdots,\mathbf{y}_{m}\}$ are given. 
Each graph $\mathcal{G}_{i}=(\mathbf{A}_{i},\mathbf{X}_{i})$ is described by an adjacency matrix $\mathbf{A}_{i}\in \mathbb{R}^{m_{i} \times m_{i}}$ and a node feature matrix $\mathbf{X}_{i} \in \mathbb{R}^{m_{i} \times d}$, 
where ${m}_{i}$ denotes the number of nodes in $\mathcal{G}_{i}$ and $d$ is the dimension of node feature. Then, according to label $\mathbf{y}$, we split $G$ into $\{(G^{train},\mathbf{y}^{train})\}$ and $\{(G^{test},\mathbf{y}^{test})\}$ as train set and test set respectively. 
Notice that $\mathbf {y}^{train}$ and $\mathbf{y}^{test}$ must have no common classes. The target is to learn a classifier from $\{(G^{train},\mathbf{y}^{train})\}$ (i.e., meta-train), that can be generalized to making prediction for classes in $\mathbf{y}^{test}$ given only a few labeled graphs (i.e., meta-test).

In the meta-train and meta-test phases, 
we utilize the episodic method of task learning~\citep{MAML} in the conventional few-shot learning.
Taking the meta-train phase as an example, we randomly sample meta-tasks, which contain a support set $D_{sup}=\{ (G_{sup},\mathbf{y}_{sup}) \}$ and a query set $D_{qry}=\{(G_{qry},\mathbf{y}_{qry}) \}$.
Given labeled support data, our goal is predicting the labels of query data. 
Generally, there are $N$ classes in the support set, and each class has $K$ samples. This is the $N$-way $K$-shot graph classification problem we aim to solve.
For each meta-task, its support set and query set have the same classes, but the number of samples in each class is not necessarily the same. 

\subsection{The Proposed Framework}\label{ssec:framework}
\begin{figure*}
	\centering
	\includegraphics[width=0.8\textwidth]{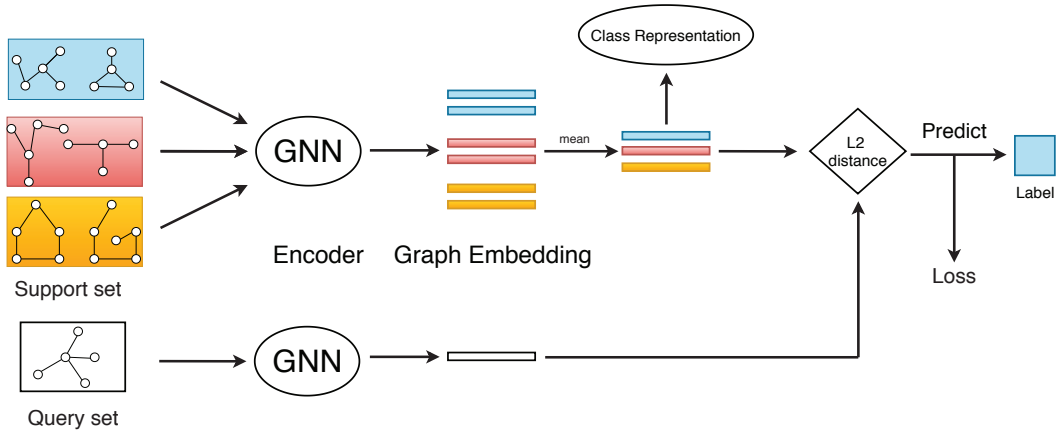}
	\caption{Illustration of metric-based meta-learning for few-shot graph classification.}
	\vspace{-15pt}
	\label{fig:framework}
\end{figure*}

Inspired by the metric-based meta-learning method in the field of computer vision~\citep{simpleshot}, the framework (see Figure~\ref{fig:framework}) learns a GNN as an encoder to project a graph (i.e., $\mathcal{G}_{i}$) into a latent representation $\mathbf{h}_i = f(\mathcal{G}_{i} | \theta)$, and makes prediction based on the nearest neighbor principle by calculating distance between the query graph and the graphs in the support set according to their latent representations.
For a $N$-way $K$-shot meta-task, 
by feeding the support set $D_{sup}=\{ (G_{sup},\mathbf{y}_{sup}) \}$ through the encoder, we obtain their latent representations, $\mathbf{h}_{G_{sup}}^{i,n}(i\in[1,K], n\in[1,N])$. From the representations of the graphs in each class, we obtain a class centroid $\mathbf{c}_n$, which is formulated as:
\begin{equation}
\label{eq:class representation}
\mathbf{c}_{n} = \frac{1}{K} \sum_{i=1}^{K}\mathbf{h}_{G_{sup}}^{i,n}.
\end{equation} 
Similarly, we extract the representation of query graphs, $\mathbf{h}_{G_{qry}}^{i}$($i\in[1,Q]$, where Q is the number of query set) of all samples.

\paragraph{Meta-train.}
We measure the distance between the the query graph and each class centroid. The nearest neighbor classification method is utilized to predict the label of each query graph:
\begin{equation}
\begin{split}
\label{qr:predict}
\hat{y}_{qry}^{i} = \argmin_{n} d\left(\mathbf{c}_{n}, \mathbf{h}_{G_{qry}}^{i}\right),
\end{split}
\end{equation}
where $d(\cdot)$ is a distance metric such as the Euclidean distance.
The parameters of the encoder are learned by optimizing a classification loss over the query graphs, which is:
\begin{equation}
\label{eq:loss}
\argmin_{\theta}\sum_{i} \mathcal{L}\left(
    \hat{y}_{qry}^{i}, y_{qry}^i
\right),
\end{equation}
where $y_{qry}^i$ is the label of the query graph $i$ and $\mathcal{L}$ is selected to be the widely used cross-entropy loss.

\paragraph{Meta-test.}
Given a graph $\mathcal{G}_{i}$ from a meta-task in the meta-test phase, instead of a direct usage of the extracted representation $\mathbf{h}_i = f(\mathcal{G}_{i} | \theta)$. We consider a representation transformation to facilitate the generalization to new classes by centering and scaling, inspired by the superior performance of SimpleShot~\citep{simpleshot}.
1) Centering. As the encoder is trained over graphs from the training graphs, it may induce some bias into the representation of graphs from unseen classes. Therefore, we subtract the mean representation over all graphs occurred in meta-train to mitigate such representation drift. Formally,
\begin{equation}
\label{eq:mean}
\bar{\mathbf{h}}_i = f\left(
    \mathcal{G}_{i} | \theta
\right) - \frac{1}{|\mathcal{G}_{train}|} \sum_{\mathcal{G}_j \in \mathcal{G}_{train}} f(\mathcal{G}_{j} | \theta).
\end{equation}
2) Scaling. We further conduct L2-normalization over $\bar{\mathbf{h}}_i$ so that the range of distance across different classes is close to each other. That is, we use
$\hat{\mathbf{h}}_i \leftarrow \frac{\bar{\mathbf{h}}_i}{\|\bar{\mathbf{h}}_i\|_2}$ to calculate class centroid and graph distance in meta-test.

In particular, we choose GIN as the backbone of the encoder, which is one of the state-of-the-art models for graph classification. Upon the GIN model, we then detail our design for considering \textit{global structure} and \textit{local structure} when learn the latent representation $\mathbf{h}_i$. It should be noted that the schema can also be applied to other GNN graph classification models such as DiffPool~\citep{diffpool}. In the following, the omit the subscript for the briefness of the notations.

\subsection{Global structure and Local structure}

\begin{figure*}
    \centering
    \subfigure[Considering \textit{global structure} in the GIN encoder.]{\includegraphics[width=0.65\textwidth, clip]{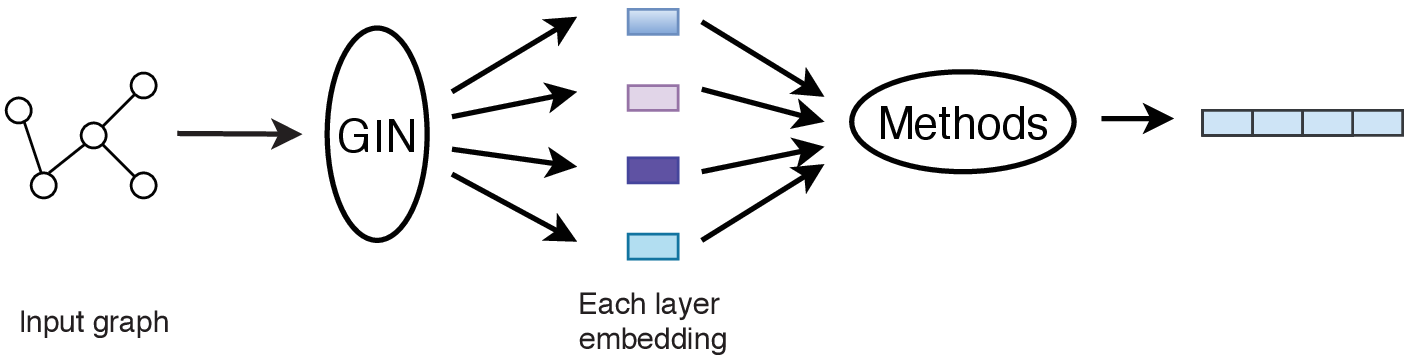}}
    \subfigure[Considering \textit{local structure} in the GIN encoder.]{\includegraphics[width=0.72\textwidth, clip]{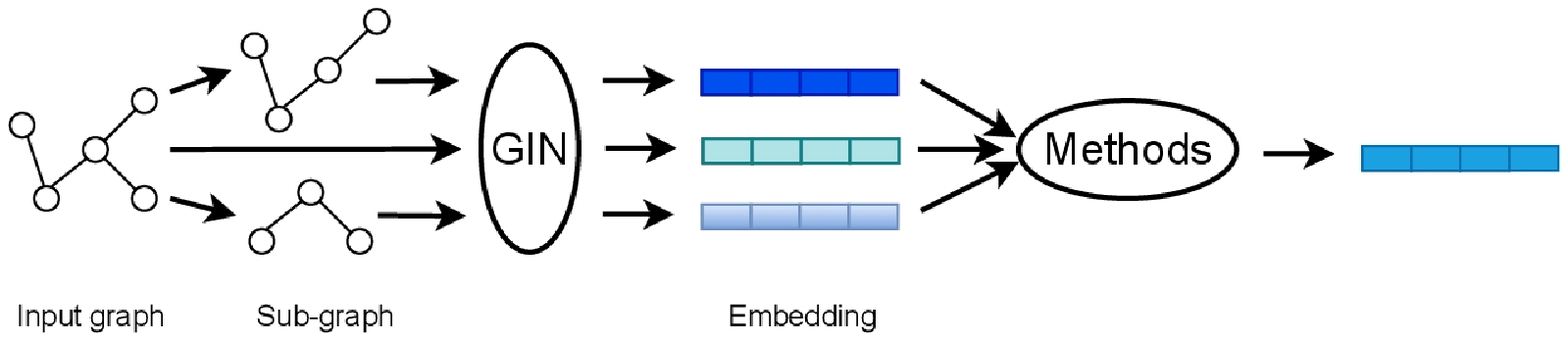}}
    \vspace{-0.3cm}
    \caption{The consideration of graph \textit{local structure} and \textit{global structure} in the GIN encoder.}
    \label{fig:local-gloabl}
    \vspace{-0.5cm}

\end{figure*}

As shown in Figure~\ref{fig:local-gloabl}, we devise two modules upon GIN to encode the \textit{global structure} and \textit{local structure} and enhance the graph representation.

\subsubsection{Global structure}

Generally, if there are $l$ layers in GIN, each node will aggregate $l$-hop neighbors, and the graph representation at each layer $\{\mathbf{h}^{1},\mathbf{h}^{2},\cdots,\mathbf{h}^{l}\}$ are realized by a READOUT function such as average pooling, which aggregates the node representations. By concatenating the representation at each layer, 
we obtain the entire graph representation. Formally,
\begin{align}\label{eq:r-g-con}
    \mathbf{h}_{\mathcal {G}}=con(\mathbf{h}^{1}, \mathbf{h}^{2}, \cdots, \mathbf{h}^{l}).
\end{align}
Intuitively, for graphs with different \textit{global structure}, the representation at different layers should be highlighted differently rather than simply treated equally.
For instance, the representation of $\mathbf{h}_{\mathcal {G}}$ should highlight the representation from shallow layers to avoid the impact of over-smoothing.
In order to make full use of the \textit{global structure}, we use a \textit{global structure} attention to learn $\{w_1, w_2, \cdots, w_l\}$ to model the importance of different layers:
\begin{equation}
\label{eq:r-g}
\mathbf{h}_{\mathcal {G}} = con\left( w_1\times \mathbf{h}^{1}, w_2\times \mathbf{h}^{2}, \cdots, w_l\times \mathbf{h}^{l}\right).
\end{equation}
In this way, the representation from a layer adaptively contributes to the representation of different graphs. In Section~\ref{sssec:specific}, we detail the implementation of the \textit{global structure} attention. Note that we follow the previous GNN models for graph classification, which typically uses concatenation to aggregate the representations across difference layers.

\subsubsection{Local structure}
We believe that the attribute characteristics of graph data depend on the substructure characteristics. In other words, the local substructure in the graph plays a decisive role in the label prediction of the entire graph. 
In many domains, the crucial substructure can be recognized according to domain knowledge such as the scaffold of molecular\footnote{In case such domain knowledge is not available, we simply divide the entire graph into two substructures.}.

Let $\mathcal{G}_{subs} = \left\{\mathcal{S}_{1}, \cdots, \mathcal{S}_{n}\right\}$ denote the crucial substructures extracted from graph $\mathcal {G}$. 
For each substructure, e.g., $\mathcal{S}_{1}$, we use the same GIN encoder to calculate its latent representation at each layer and obtain the substructure representation $\mathbf{h}_{\mathcal{S}_{1}}$ according to Equation~\ref{eq:r-g}\footnote{Note that we can also calculate the representation according to Equation~\ref{eq:r-g-con} if ignoring the global structure attention.}.
Considering that substructure can contribute unequally to the prediction of different graphs, we devise a \textit{local structure} attention to model the weight of substructure and obtain the final representation of the graph $\mathbf{h}$. Formally, 
\begin{equation}
\label{eq:r-l}
\mathbf{h} = r_{0}\times \mathbf{h}_{\mathcal{G}} + r_1\times \mathbf{h}_{\mathcal{S}_1}, \cdots, r_n\times \mathbf{h}_{\mathcal{S}_n},
\end{equation}
$\{r_0, r_1, \cdots, r_n\}$ is calculated by an attention model (see Section~\ref{sssec:specific}). If not otherwise specified, we aggregate the entire graph representation $\mathbf{h}_{\mathcal{G}}$ and the substructure representations through mean-pooling, which is a common setting in the previous usage of attention model~\citep{diffpool} and is sufficient for distilling the important signals for making classification (see detailed results in Section~\ref{sssec:pooling}).


\subsubsection{Attention models}\label{sssec:specific}

For both the \textit{global structure} attention and the \textit{local structure} attention, the target is to learn a group of weights (i.e., $w$ and $r$) with a set of representations as inputs, and aggregate the representations according to the learned weights. Considering that graphs changes dramatically across different domains, we try five different attention models to calculate the weights. Without loss of generality, we elaborate the attention models with the \textit{global structure} attention as an example where the inputs are $\mathcal{H} = \{
    \mathbf{h}^{1}, \mathbf{h}^{2}, \cdots, \mathbf{h}^{l}
\}$.

\paragraph{Learned-weight.}

A straightforward implementation is treating the target weights as additional model parameters directly which are updated as normal parameters during model training. Formally, we introduce a learnable weight vector $\mathbf{w}=[w_1, w_2, \cdots, w_l]$. As the weights are shared across different graphs, this method is suitable for scenarios where the \textit{global structure} properties are consistent across graphs, especially the ones without enough labeled data since the minimal number of additional model parameters.

\paragraph{Vanilla attention.}

Inspired by the success of the vanilla attention model~\citep{cwj}, it can be a better choice for implementing our attentions, which is formulated as:

\begin{equation}
\begin{split}
w_{j}=\frac{\exp(e_{j})}{\sum_{k \leq |\mathcal{H}|}\exp(e_{k})},~\text{where}~e_{j}=\mathbf{c}^{T}tanh(\mathbf{W}\mathbf{h}^j+\mathbf{b}),
\end{split}
\label{eq:attention}
\end{equation}
where $\mathbf{c}$, $\mathbf{W}$, and $\mathbf{b}$ are model parameters to be learned.

\paragraph{Self-attention.}

Furthermore, to distinguish the different impact among representations, we further referenced the multi-head self-attention mechanism~\citep{attention}. We follow the standard implementation of multi-head self-attention~\citep{attention}
where the input representations are fed into the model as a sequence. Note that the learned weights have been absorbed into the output representations.

\paragraph{MLP.}
Similarly, we can embed the weight into a Multi-Layer Perceptron (MLP) to learn the aggregation of the input representations, which is formulated as:

\begin{equation}
\label{eq:mlp}
\mathbf{h}^{(k)} =  {\rm ReLU}\big( \mathbf{W}^{(k - 1)}\mathbf{h}^{(k - 1)} + \mathbf{b}^{(k - 1)} \big),
\end{equation}
where $k$ is the number of MLP layers, $\mathbf{h}^{(0)} = con(\mathcal{H})$ is a concatenation of the input representations, and $\mathbf{h}^{(k)}$ denote the output representation. 

\paragraph{Transformer.}

To further enhance the representation ability, we can also implement the attentions by a Transformer layer, which is combination of multi-head self-attention and MLP. 
We refer the original paper~\citep{attention} for the detailed formulation of a Transformer layer.

\subsubsection{Our framework and variants}

On the basis of these five different calculation methods, we consider the characteristics of both \textit{global structure} and \textit{local structure},
and implement our framework with \textit{global structure} attention and \textit{local structure} attention simultaneously, named SMF-GIN. In particular, SMF-GIN first calculates the representation of the entire graph and each substructure according to Equation~\ref{eq:r-g}, and then calculates the final representation for making classification according to Equation~\ref{eq:r-l}.

Moreover, we also study three variants of SMF-GIN:
\begin{itemize}[leftmargin=*]
    \item \textbf{SMF-GIN-$G$,} which only includes \textit{global structure} attention, i.e., taking the output of Equation~\ref{eq:r-g} as the final representation.
    \item \textbf{SMF-GIN-$L$,} which only considers the characteristics of \textit{local structure}. As compared to SMF-GIN, SMF-GIN-$L$ calculates the representation of the entire graph and each substructure according to Equation~\ref{eq:r-g-con}.
    \item \textbf{SMF-GIN-$E$.} Lastly, we ensemble SMF-GIN-$G$ and SMF-GIN-$L$, named SMF-GIN-$E$, through a voting mechanism. In the meta-test phase, SMF-GIN-$E$ averagely aggregates the prediction of a query graph (i.e., distance to each class centroid) from SMF-GIN-$G$ and SMF-GIN-$L$.
\end{itemize}

\section{Experiments}
In this section, we will introduce experiment datasets, baselines and details of our methods for implementation.
In our experiments, we focus on four research questions: 
(1) Could our framework SMF-GIN, considering the characteristics of both \textit{global structure} and \textit{local structure}, achieve better results?
(2) To what extent, the consideration of \textit{global structure} and \textit{local structure} improve the classification results?
(3) What are the specific performances of different attention models in SMF-GIN-$G$ and SMF-GIN-$L$?
(4) What is the impact of hyper-parameters on SMF-GIN-$G$ and SMF-GIN-$L$? 
In response to these four questions, we conduct experiments and answer these questions in the analysis of experimental results.

\subsection{Experiment Settings}
\paragraph{Datasets.}
We conduct experiments on two datasets: the multi-class Chembl dataset and public dataset TRIANGLES. Detailed information about these two datasets illustrated in the Table \ref{tab:datasets}.

\textbf{Chembl.}
In order to promote the few-shot graph classification investigation and evaluation, we constructed a general large-scale multi-class graph classification benchmark dataset from the Chembl chemical molecule database~\citep{MoleculeNet},
which contains approximately 1.88 million molecules and 12,482 molecular targets. Specifically, select the target type as protein, the target confidence score is higher than 0, and each molecule contains only one target. For better training, select the target with more than 500 corresponding molecules.
Take the final 273 molecular targets as the categories of our dataset, and randomly select 500 samples from each category as the final Chembl dataset, a total of 136500 molecules.

\textbf{TRIANGLES.}
This dataset contains ten different graph classes numbered from 1 to 10, corresponding to the number of triangles in each graph in the dataset.
Each class of this dataset contains 201 graphs, for a total of 2010 graphs.

\begin{table}[]
\caption{Datasets statistics and splits.}
\label{tab:datasets}
\centering
\resizebox{0.45\textwidth}{8mm}{%
\begin{tabular}{ccccc}
\hline
\multirow{2}{*}{Dataset Name} & \multicolumn{3}{c}{\#Classes} & \multirow{2}{*}{\#Classes Per Graph} \\ \cline{2-4}
                              & \#Train   & \#Validation  & \#Test  &                                    \\ \hline
Chembl                        & 219     & 27          & 27    & 500                                \\
TRIANGLES                     & 6       & -           & 4     & 201                                \\ \hline
\end{tabular}
}

\end{table}

\paragraph{Baselines}
For the N-way K-shot graph classification task, we utilize 5 state-of-the-art baseline methods to compare with our methods: Pre-GNN, GIN, MAML, Supclass and SimpleShot. 
Detailed information about these 5 types of baselines is included:

\begin{itemize}[leftmargin=*]
    \item \textbf{Pre-GNN~\citep{pre-GNN}:}
This method pre-trains the expressive GNN of a single node and an entire graph so that GNN can learn useful local and global representations at the same time. We utilized three pre-trained models: Pre-context, Pre-masking, and Pre-infomax. In the test phase, N-way K-shot tasks were used to fine-tune these three pre-trained models.

\item \textbf{GIN~\citep{GIN}:}
This baseline method, follows a neighborhood aggregation strategy, where we iteratively update the representation of a node by aggregating representations of its neighbors. In the training phase, we utilize all categories' train set graphs to train the encoder and classification layer. In the test phase, we discard the trained classification layer and retain the encoder. Then use the N-way K-shot task to fine-tune the encoder and the new classification layer.

\item \textbf{MAML~\citep{maml-GNN}:} 
We apply gradient-based meta-learning method~\citep{MAML} to GNNs model. More specifically, we leverage GIN~\citep{GIN} as the graph encoder backbone and meta-learning as a training paradigm to capture task-specific knowledge rapidly in the graph classification tasks and transfer them to new tasks.

\item \textbf{Supclass~\citep{supclass}:} 
This method was proposed based method customized for the few-shot graph classification. On the training stage, they compute prototype graphs from each class, then they cluster the prototype graphs to produce superclasses. After that, they predict the origin class and superclass of each graph. On the test stage, they only update the classifier based on the classification of origin classes.

\item \textbf{SimpleShot~\citep{simpleshot}:}
We apply the metric-based meta-learning method to the GNNs model without considering graph structure characteristics. Specifically, we utilize GIN as the encoder to find the class representation in support set, and then build a metric space shared with the query set. Perform transformation on the graph representation, and finally utilize the nearest neighbor principle for N-way K-shot classification.
\end{itemize}



\begin{table}[]
\caption{Accuracies with a standard deviation of baseline methods and our framework. We tested 1000 and 500 N-way-K-shot tasks on Chembl and TRIANGLES respectively. The bold black numbers denote the best results we get.}
\label{tab:results}
\centering
\resizebox{0.48\textwidth}{30mm}{%
\begin{tabular}{lcc}
\hline
\multicolumn{1}{c}{\multirow{2}{*}{Methods}} & Chembl               & TRIANGLES           \\ \cline{2-3} 
\multicolumn{1}{c}{}                         & 5-way 5-shot         & 3-way 5-shot        \\ \hline
Pre-context                                  & $45.75\pm0.61$          & -                   \\
Pre-masking                                  & $45.95\pm0.62$          & -                   \\
Pre-infomax                                  & $44.09\pm0.60$          & -                   \\
GIN                                          & $64.85\pm0.57$          & $65.31\pm0.98$         \\
MAML                                         & $61.69\pm0.57$          & $72.29\pm0.97$         \\
Supclass                                     & $59.17\pm0.56$          & $71.43\pm0.93$          \\
SimpleShot                                   & $73.08\pm0.5$4          & $75.56\pm0.86$          \\ \hline
SMF-GIN-$G$                                  & $74.37\pm0.54$          & $78.16\pm0.74$         \\
SMF-GIN-$L$                                  & $74.10\pm0.54$          & $78.27\pm0.82$           \\
SMF-GIN                                      & $71.12\pm0.56$          & $75.57\pm0.85$          \\
SMF-GIN-$E$(self)                            & $74.36\pm0.55$          & $79.06\pm0.76$          \\
SMF-GIN-$E$(best)                            & \textbf{75.27$\pm$0.53} & 79.85$\pm$0.73 \\ \hline
\end{tabular}
}
\end{table}

\paragraph{Experimental implementation details.}
On account of the number of categories between the chembl dataset and the TRIANGLES dataset is too different, we perform 5-way 5-shot classification task on the Chembl dataset and 3-way 5-shot classification task on the TRIANGLES dataset, where the number of samples in each class in the query set is 15. The split of the train set, validation set and test set is illustrated in Table \ref{tab:datasets}. 
We utilize the 5-layer GIN model as the encoder (each layer contains 2 layers of MLP). Among them, we CONCAT the graph representation in the last 4 layers as the real graph representation, because the first layer is the original node feature.
We set the dimension of the hidden layer to 64 and the learning rate to 0.001. During the training process, a total of 700 iterations are performed, and validation is performed every 20 epochs. 
The model with the best verification results is selected for testing to obtain the final classification result.
We implement the proposed solution on Pre-GNN\footnote{\url{https://github.com/snap-stanford/pretrain-gnns}.} through PyTorch 1.0.1. As to the substructure, we adopt the scaffold of the molecular in the Chembl dataset, and a random split of the graph in the TRIANGLES dataset.

\subsection{Analysis of results}
Our experimental results are illustrated in Table \ref{tab:results}. Pre-GNN is not suitable for the TRIANGLES dataset, because this model is designed to process chemical molecule and biological protein graph so that it could only process the Chembl dataset. We can observe that three methods of Pre-GNN perform badly on the N-way K-shot classification task. Here, we believe that there is a great difference between the dataset used in Pre-GNN and the graph structure of the dataset we finally test, so even fine-tuning the model on the N-way K-shot task has not achieved good results.

Regarding the utilization of MAML in GNN to solve the few-shot graph classification task, in the Chembl dataset, MAML is not as effective as simply using GIN, while in the TRIANGLES dataset, the result is just the opposite.
The reason for this phenomenon is the different structure of the graphs,
the differences between the Chembl dataset categories are large, whereas the TRIANGLES dataset represents only the number of triangles in the graph, with a small structural difference.
This is consistent with the general interpretation of MAML: the generalization ability is poor when there are large differences in sample data.
In both Chembl and TRIANGLES datasets, the performance of MAML is higher than Supclass, which is also consistent with the result in~\citep{maml-GNN}. 

In all models, without considering structural characteristics, the best performance is SimpleShot, the application of the metric-based meta-learning method to the GNNs model.
It can be seen that it is effective to share a metric space between the graph data for N-way K-shot classification task.
For our framework: SMF-GIN, the performance is superior to that of baselines except SimpleShot because package the characteristics of both \textit{global structure} and \textit{local structure} into an end-to-end solution would interfere with the training process.
Therefore, the answer to our first question is yes.

For the second question, we also implement SMF-GIN variants: SMF-GIN-$G$ and SMF-GIN-$L$, which includes \textit{global structure} attention and \textit{local structure} attention, respectively. we can make the following observation: compared with the baseline and SMF-GIN, SMF-GIN-$G$ and SMF-GIN-$L$ both achieve better performance, which shows that it is easier to capture graph structural characteristics when considering \textit{local structure} or \textit{global structure} separately.

At the end of the previous section, we ensemble SMF-GIN-$G$ and SMF-GIN-$L$, named SMF-GIN-$E$ through a voting mechanism.
Here we consider two ways of aggregating: SMF-GIN-$E$(self) and SMF-GIN-$E$(best).
In particularly, SMF-GIN-$E$(self) refers to the results of using Self-attention mechanism in both SMF-GIN-$G$ and SMF-GIN-$L$, while SMF-GIN-$E$(best) refers to the results of adopting the optimal method in both SMF-GIN-$G$ and SMF-GIN-$L$.
As shown in Table \ref{tab:results}, SMF-GIN-$E$(best) gets the best result and SMF-GIN-$E$(self) gets the second best result. This shows that it is important to make full use of the specific structure of graph in the few-shot graph classification task.

\begin{figure}[t]
    \centering
    \includegraphics[width=.45\textwidth]{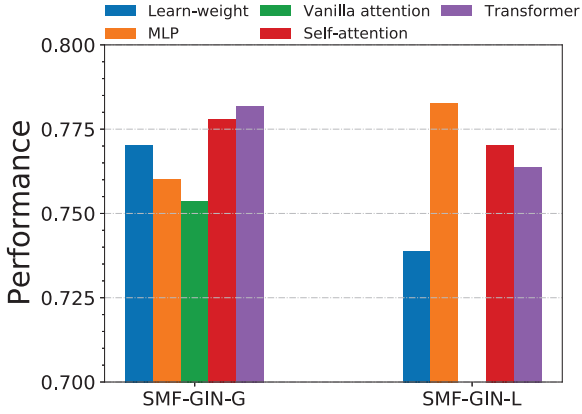}

    \caption{Impact of five attention models in SMF-GIN-$G$ and SMF-GIN-$L$.}
    \label{fig:ways}
    \vspace{-0.7cm}
\end{figure}

\subsection{Research on attention models and hyper-parameters}
In the previous section, we put forward five different attention models to both the \textit{global structure} attention and the \textit{local structure} attention. In this section, we take the TRIANGLES dataset as an example and compare the performance difference of five attention models on SMF-GIN-$G$ and SMF-GIN-$L$ through experiments.
For the self-attention mechanism, we have studied the influence of some hyper-parameters and pooling strategies on the final result.

\subsubsection{Five attention models}
As illustrated in Figure \ref{fig:ways}, we can see the performance of these different attention models on SMF-GIN-$G$ and SMF-GIN-$L$. 

In SMF-GIN-$G$, the vanilla attention omits the interactions across different layers' representation, so the generalization ability for each layer representation is relatively poor.
Simply using learned-weight or MLP layers to accommodate each layer's representation is not as effective as self-attention and Transformer mechanisms.
Because shallow representation (e.g., at the first layer) and deep representation (e.g., at the last layer) have different importance for different graphs, and self-attention and Transformer mechanisms can further enhance the representation ability and distinguish the different impact among representations well.

In SMF-GIN-$L$, due to the poor generalization ability of the vanilla attention mechanism for representations of the entire graph and substructures, the result is far lower than the other attention models, and its performance cannot be illustrated in Figure \ref{fig:ways}. Simply using learned-weight is difficult to fit different substructures, more parameters are conducive to the entire graph and substructures' representation in SMF-GIN-$L$, so that the effectiveness of MLP method is higher than self-attention and Transformer mechanisms. 

\subsubsection{The effect of layer depth}
\begin{figure}[t]
    \centering
    \subfigure[SMF-GIN-$G$]{\includegraphics[width=.238\textwidth, clip]{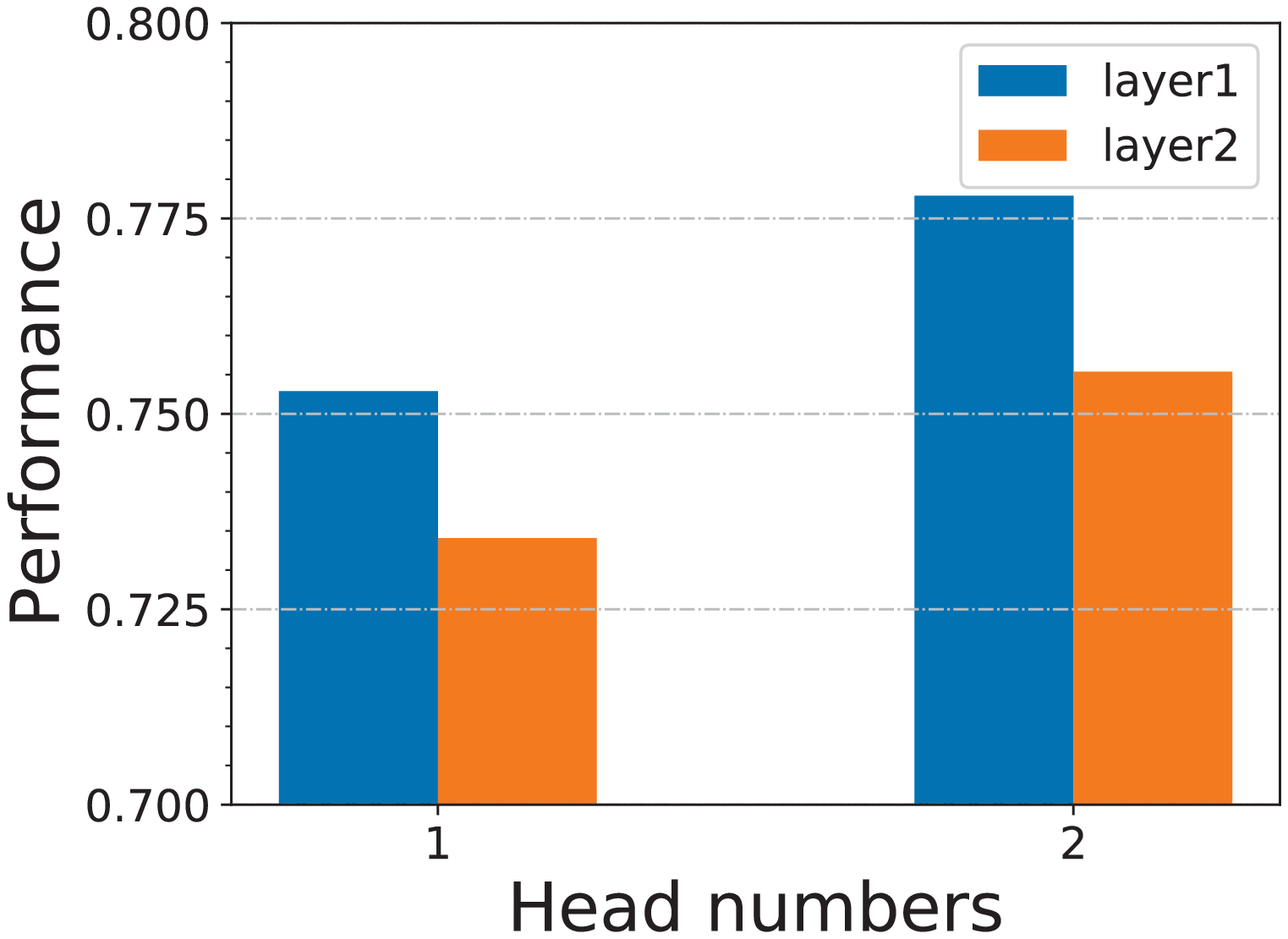}}
    \subfigure[SMF-GIN-$L$]{\includegraphics[width=.238\textwidth, clip]{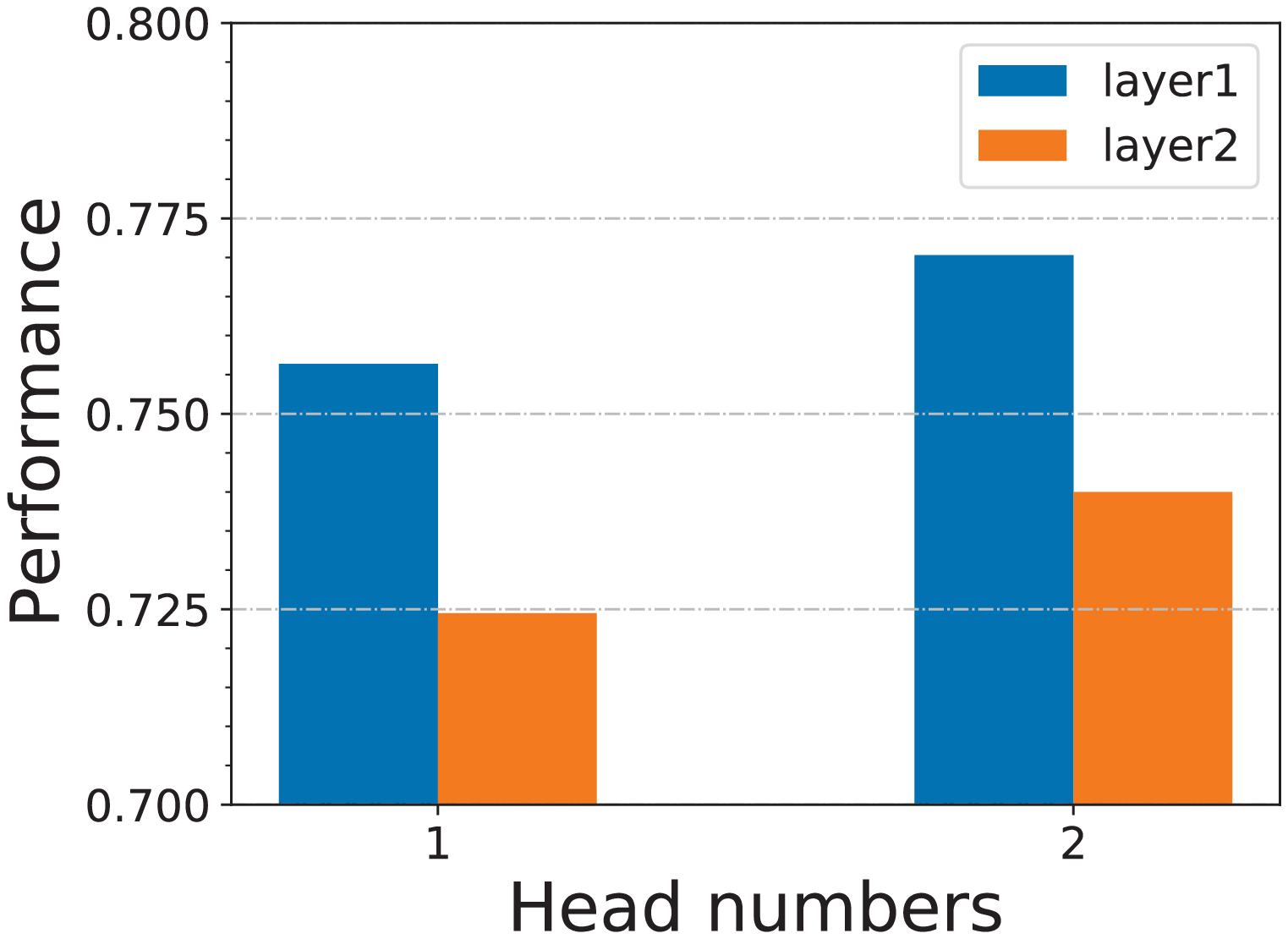}}    
    
    \caption{Impact of layer depth in (a) SMF-GIN-$G$ and (b) SMF-GIN-$L$.}
    \label{fig:layer}
    \vspace{-0.4cm}

\end{figure}

As illustrated in Figure \ref{fig:layer}, we compare layer=1 and layer=2 in self-attention mechanism on head=$\{1,2\}$ in SMF-GIN-$G$ and SMF-GIN-$L$ respectively. 
According to Figure \ref{fig:layer}, we observe that the effect of layer depth on model performance is the same in SMF-GIN-$G$ and SMF-GIN-$L$, 
When layer=1, the performance both are higher than layer=2 whether the head number is $1$ or $2$.
The reason is that in the self-attention mechanism, representations of the entire graph and substructures and the encoder each layer representations needn't deep aggregation, and shallow layer can be expressed well.

\begin{figure}[t]
    \centering
    \includegraphics[width=.35\textwidth]{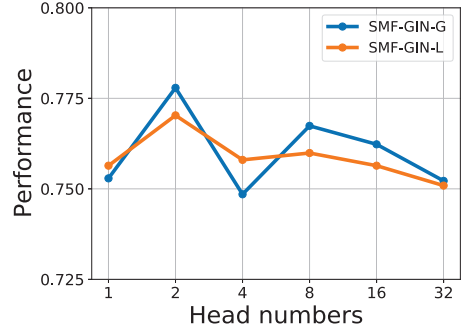}

    \caption{Impact of head numbers in SMF-GIN-$G$ and SMF-GIN-$L$.}
    \label{fig:head}
    \vspace{-0.7cm}
\end{figure}

\subsubsection{The effect of head numbers}
Comparing the impact of layer depth on performance in Figure \ref{fig:layer}, we found that when the head numbers increases, performance in SMF-GIN-$G$ and SMF-GIN-$L$ is greatly improved. Therefore, we fixed layer=1, tried to change head numbers, and observed its performance trend. As illustrated in Figure \ref{fig:head}, with the increase of head numbers, there is the same change trend in SMF-GIN-$G$ and SMF-GIN-$L$, and the performance reaches its peak when head=2. Overall, when head$\leq$8, the fluctuation of SMF-GIN-$G$ is larger than SMF-GIN-$L$, indicating the different layers' representation in SMF-GIN-$G$ are more sensitive to head numbers. When head$>$8, the performance drops slowly, indicating that larger head numbers are harmful to this model.

\subsubsection{Impact of pooling strategies}\label{sssec:pooling}
\begin{figure}[t]
    \centering
    \includegraphics[width=.35\textwidth]{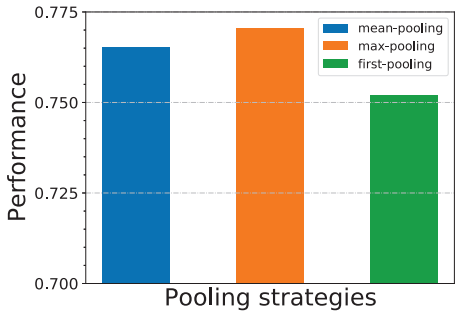}

    \caption{Impact of Pooling strategies in SMF-GIN-$L$.}
    \label{fig:pooling}
    \vspace{-0.7cm}
\end{figure}

Recall that we take mean-pooling as a default choice to aggregate the entire graph representation and substructure representations (c.f., Equation~\ref{eq:r-l}). We then investigate the impact of different pooling strategies on the final prediction results.  
We fix previously analyzed hyper-parameters as the best choice, that is $\{layer=1, head=2\}$, and selected max-pooling, mean-pooling and first-pooling, where first-pooling represents the first dimension of the output directly, that's the entire graph's representation $\mathbf{h}_{\mathcal{G}}$. 
As illustrated in Figure \ref{fig:pooling}, we can see that max-pooling has the best performance, mean-pooling is the second, and first-pooling is the worst. 
Because max-pooling is more like making the best feature selection compared to others, selecting the representation with higher classification recognition, providing non-linearity and the highest performance naturally.

\section{Related Work}
In this section, we briefly introduce the relevant research lines of our work: graph classification, few-shot learning and graph few-shot learning.
\subsection{Graph classification}
Graph Neural Networks (GNNs) were first introduced in~\citep{gori2005}. This algorithm follows the neighborhood aggregation framework, computes the representation vector of the node by the recursive aggregation and transformation of adjacent nodes feature vector. As the main branch in recent years, GNNs have been successfully applied to graph classification. ~\citep{GCN} proposed Graph Convolutional Neural Network (GCN) and achieved inspiring results based on feature aggregation from neighborhood nodes. ~\citep{GAT} introduced attention mechanism for graph convolutional operations (GAT). ~\citep{diffpool} proposed DIFFPOOL, a differentiable graph pooling module that can generate hierarchical representations of graphs and can be combined with various graph neural network architectures in an end-to-end fashion. ~\citep{GIN} developed a simple Graph Isomorphism Network (GIN) that is provably the most expressive among the class of GNNs and is as powerful as the Weisfeiler-Lehman graph isomorphism test. 

When these graph classification methods encounter novel classes, the existing GNNs models always need to re-learn their parameters to incorporate the new information, and if there are few samples of data in each new class, the performance of the GNNs model will suffer a catastrophic decline. However, our framework uses a metric-based meta-learning method with GIN as the model, it can achieve good results when there are few novel class data samples.

\subsection{Few-shot learning}
Few-shot learning in the computer vision community was first introduced by~\citep{one-shot}. 
Its goal is to generalize a model to a new task rapidly with only a small number of samples after learning a large amount of data with a certain category.
Various learning algorithms have been proposed in the image domain, and recent related works can be classified into two categories: gradient-based methods and metric-based methods.
\paragraph{Gradient-based methods}
These methods aim to learn a better initialization of model parameters that can be updated by a few gradient steps in future tasks to fast adapt to new tasks~\citep{MAML}~\citep{transfer}, or train parameter generator for task-specific classifier~\citep{ruse}.
\paragraph{Metric-based methods}
The purpose of these methods is to learn a metric space from the training task. When a new task appears, this metric space is shared with the new task~\citep{Vinyals}~\citep{snell}.

These methods have achieved inspiring results in image classification tasks, but they have not been well explored in graph data.
In contrast, our framework adopts the metric-based method and performs well in the graph classification task.

\subsection{Graph few-shot learning}
Based on this challenge, recently many scholars have carried out researches on few-shot learning in graph data. ~\citep{SRGCN} solved the sparsity problem by adding regularization in the loss function of the graph neural network in the few-shot node classification task.
~\citep{GFL} proposed a graph few-shot learning (GFL) algorithm based on the prior knowledge of auxiliary graphs, which transferred the knowledge learned from the auxiliary graphs to the new target graphs to improve the effectiveness of semi-supervised node classification.
~\citep{meta-graph} adopted the traditional meta-learning method to optimize the shared parameter initialization of the local link prediction model. Currently, these few-shot learning methods are widely used in node classification and link prediction tasks, but not suitable for graph classification tasks. 
~\citep{maml-GNN} adopted gradient-based meta-learning methods to deal with the few-shot graph classification task, updated the initialization parameters of the model according to the different meta-task in order to fast adapt to new tasks. ~\citep{supclass} proposes a graph distance measure in the spectral domain, which validates the merit of considering graph structure.

However, these methods focus on algorithms and models without utilizing the characteristics of graph data structure well.
In order to make up for this shortcoming, we put forward our framework under the premise of considering the structure of graph data explicitly.

\section{Conclusion and Future Works}
This paper highlights the structure-enhanced meta-learning for few-shot graph classification.
We proposed a framework, named SMF-GIN, for the N-way K-shot graph classification problem. It uses the GIN as an encoder to learn graph representations, where two attention mechanisms are designed to encode the \textit{global structure} and \textit{local-structure}.
Five attention models are tested under the proposed framework. 
Extensive experimental analysis shows that for few-shot learning on the graph, it is more important to make full use of the structural characteristics of graphs. Meanwhile, a large-scale benchmark dataset is released to facilitate future research.
In the future, we will explore the consideration of more properties of graph structure in the few-shot setting. In addition, we will test different backbone models such as DiffPool~\citep{diffpool} and GAT~\citep{GAT}. Furthermore, we will investigate more distance measures for graphs.

\section*{Acknowledgments}
This work is supported by the National Key Research and Development Program of China (2020AAA0106000) and National Natural Science Foundation of China (U19A2079).

\bibliographystyle{cas-model2-names}

\bibliography{refs}

\end{document}